\documentclass[runningheads]{llncs}
\usepackage{graphicx}
\usepackage{subcaption}
\usepackage{graphics}
\usepackage{amsmath}
\usepackage{multirow}
\usepackage{amstext}
\usepackage[textsize=tiny]{todonotes} % todo: remove befor submission
\usepackage[export]{adjustbox}
\usepackage{mathtools}
\usepackage{breqn}
\usepackage{xcolor}
\usepackage{booktabs}
\usepackage{cite}
\usepackage{hyperref}
\usepackage{geometry}
\usepackage{enumitem}
\usepackage[nameinlink,noabbrev]{cleveref}
\begin{document}
\title{User Identification via Free Roaming Eye Tracking Data}
\titlerunning{User Identification via Free Roaming Eye Tracking Data}
\author{Rishabh Vallabh Varsha Haria\orcidID{0000-0001-8067-3608}\thanks{Corresponding author.} \and
	Amin El Abed\orcidID{0009-0008-3690-2694 } \and
	Sebastian Maneth\orcidID{0000-0001-8667-5436}}
\authorrunning{R. Haria et al.}

\institute{Department of Informatics in University of Bremen, Germany \\
	\email{haria@uni-bremen.de},
	\email{amin@uni-bremen.de},
	\email{maneth@uni-bremen.de}}
\maketitle              % typeset the header of the contribution

We present a new dataset of ``free roaming'' (FR) and ``targeted roaming'' (TR):
a pool of 41~participants is asked to walk around a university campus (FR)
or is asked to find a particular room within a library (TR). Eye movements are
recorded using a commodity wearable eye tracker (Pupil Labs Neon at 200Hz).
On this dataset we investigate the accuracy of user identification
using a previously known machine learning pipeline where a Radial Basis Function Network (RBFN) is used as classifier.
Our highest accuracies are 87.3\% for FR and 89.4\% for TR.
This should be compared to 95.3\% which is the (corresponding) highest accuracy we are aware of 
(achieved in a laboratory setting using the ``RAN'' stimulus of the~\emph{BioEye} 2015 competition dataset).
To the best of our knowledge, our results are the first that study user
identification in a non laboratory setting; such settings are often more feasible than
laboratory settings and may include further advantages.
The minimum duration of each recording is 263s for FR and 154s for TR. 
Our best accuracies are obtained when restricting to 120s and 140s for FR and TR respectively,
always cut from the end of the trajectories (both for the training and testing sessions).
If we cut the same length from the beginning, then 
accuracies are 12.2\% lower for FR and around 6.4\% lower for TR.
On the full trajectories accuracies are lower by 5\% and 52\% for FR and TR.
We also investigate the impact of including higher order velocity derivatives
(such as acceleration, jerk, or jounce).

\section{Introduction}
\label{sec:introduction}
%Eye tracking data contain information about a user’s biometric identity, gender, age, ethnicity, body weight, personality traits, emotional state, and their skills and abilities~\cite{kroger2020does}. 
%Eye tracking is a method used to gather information on the movement of a person's eyes. There are a variety of eye trackers available, ranging from expensive to more affordable options. These eye trackers are used to generate high quality datasets that serve various scientific purposes, including user identification~\cite{george2016score, kasprowski2004eye, rigas2017current, schroder2020robustness, al2022extensive}, authentication~\cite{lohr2022eye, lohr2022eye2}, personality detection~\cite{isaacowitz2005gaze, rauthmann2012eyes, hoppe2015recognition, hoppe2018eye}, disorder detection~\cite{armstrong2012eye, nilsson2016screening, billeci2017integrated}, and gender prediction~\cite{vallabh2022predicting, al2020gender}.

User identification via eye movements is a well established concept within the domain of eye tracking research~\cite{george2016score, kasprowski2004eye, rigas2017current, schroder2020robustness, al2022extensive}. 
Kasprowski and Ober's~\cite{kasprowski2004eye} seminal paper in 2004 laid the foundation for research in user identification based on eye movements. 
Since then, researchers have extensively studied the topic and established the BioEye~\cite{komogortsev2015bioeye} competition series in 2015.
It consists of two datasets: RAN (100s of recording of observing a random moving dot on a computer screen) and TEX (60s of recording of reading a poem on a computer screen). 
The winner of 2015 BioEye competition~\cite{george2016score} achieved an accuracy of 89.5\% over a single run with 153 participants. 
One important idea of their approach is to segment the data into fixations and saccades and to classify them separately. 
This paper also established the utilization of Radial Basis Function Networks (RBFN) for user identification. 
This accuracy was improved to 94.1\% in subsequent research~\cite{schroder2020robustness} by adding more features to the existing feature set developed by~\cite{george2016score}. 
In the recent study~\cite{al2022extensive} this accuracy was further increased to 96.0\% using various optimization techniques. 
The line of research continues and a new large dataset~\cite{griffith2021gazebase} was recently published using eye movements in a laboratory setting. 

One limitation of previous studies on user identification is their restriction to a laboratory environment. This approach has been criticized since it may not lead to valid theories of human behavior in natural settings~\cite{kingstone2003attention, kingstone2008cognitive}.   
However, there are significant differences in the principles guiding eye movements between looking at computer screens and engaging in dynamic real world behavior~\cite{foulsham2011and, tatler2011eye, tatler2014eye}.
A study conducted by~\cite{marius2009gaze} tracked participants' eye movements while they explored various real world environments and watched videos of these environments, providing compelling evidence for differences.
Our main contributions in this paper are: 
\begin{itemize}[noitemsep]
	\item We introduce a new dataset comprising ``free roaming" (FR) and ``targeted roaming" (TR) scenarios. In this dataset, we ask a pool of 41 users to either walk around a university campus (FR) or to locate a specific room within a library (TR).
	\item We present the first results for user identification in a non laboratory setting achieving accuracies of up to 89.4\% for TR and 87.3\% for FR dataset respectively.
	\item We compare the top features for the FR and TR datasets and find that fixation duration and saccade duration rank highest for differentiating users. 
\end{itemize}

\section{Proposed System}
\label{Proposed_system}
This section describes our user identification architecture including our dataset collection, data preprocessing, segmentation methods, feature extraction, machine learning classifiers, and the accuracy metric used in this study.

\subsection{Dataset Collection}
\begin{figure}[!ht]
	\includegraphics[width=\textwidth]{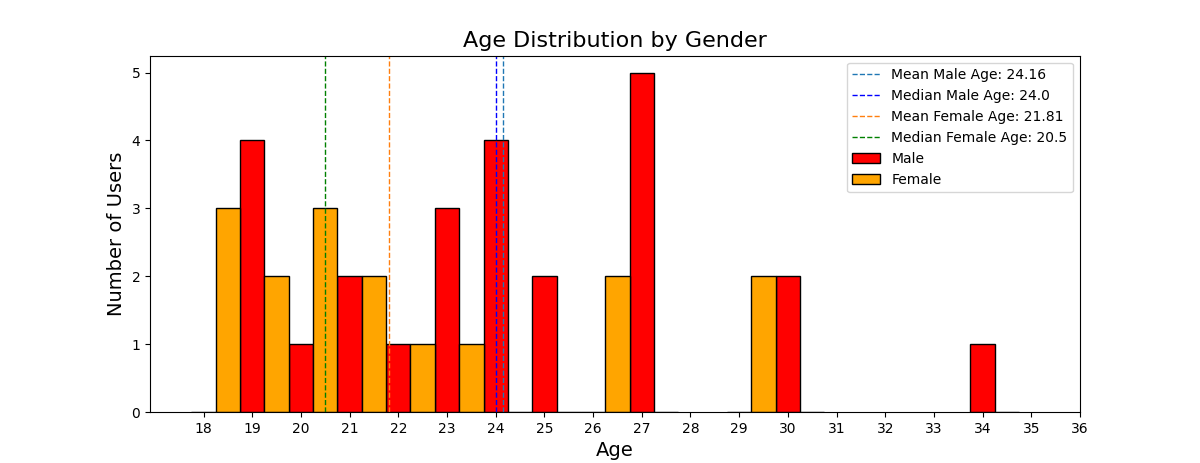}
	\caption{User distribution based on the age and gender.} \label{age_gender}
\end{figure}

The study involves 41 participants. 
The demographics of the participants are shown in Figure~\ref{age_gender}.
There are 16 females and 25 males, with the average age of males being 24, and of females being 21, and aged between 18 and 34 years. 
All participants provided written informed consent, retaining the right to withdraw from the experiment at any point. In adherence to the recommendations set forth by the Bremen University Ethics Board, only demographic variables such as age and gender are recorded in the dataset.

\begin{figure}[!ht]
	\centering
	\begin{minipage}[t]{0.4\textwidth}
		\centering
		\includegraphics[height=4cm]{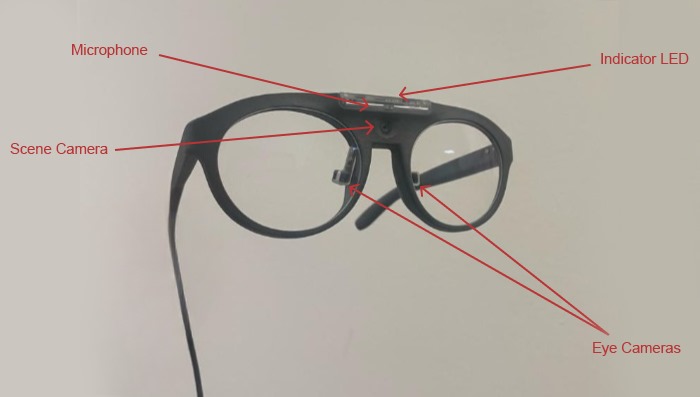}
		\caption{Pupil labs Neon eye tracker.}
		\label{fig:NEON}
	\end{minipage}
	\hfill
	\begin{minipage}[t]{0.5\textwidth}
		\centering
		\includegraphics[height=4cm]{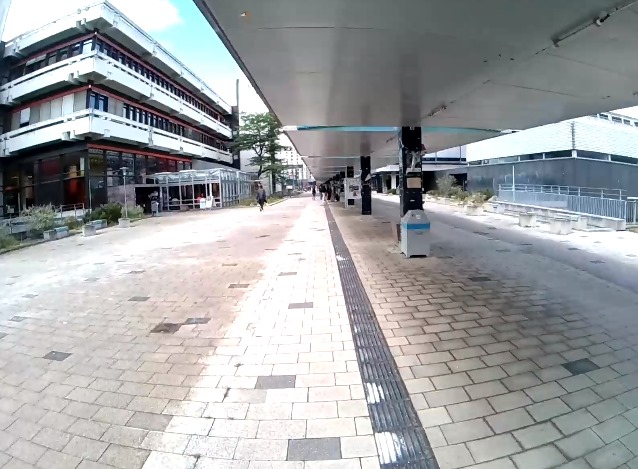}
		\caption{Boulevard at the University of Bremen.}
		\label{fig:boulevard}
	\end{minipage}
\end{figure}

%\subsubsection{Eye Tracker}
\textbf{Eye Tracker.}\quad
Gaze data from both eyes are recorded at a frequency of 200Hz using the ``Pupil Labs Neon'' eye tracker (see Figure~\ref{fig:NEON}), a commodity commercially available wearable device. 
%The eye tracker reports an gaze estimation accuracy of 1.8 degrees~\cite{baumann2023neon}.
 It utilizes two infrared eye cameras for gaze data and a scene camera for recording the surrounding environment. We delete the scene camera recordings due to anonymity requirements. 
 Calibration is not required for the eye tracker. The eye tracker comes with a mobile phone which has the Pupil Labs application and stores the recording and its associated data.

%\subsubsection{Procedure}
\textbf{Procedure.}\quad
Participants are introduced to the study and equipped with the eye tracker. Two datasets are recorded: free roaming (FR) and targeted roaming (TR).
In the free roaming scenario, participants are instructed to walk freely along the boulevard (see Figure~\ref{fig:boulevard}) at the University of Bremen. This process is conducted twice to generate two sessions (S1 and S2).
Both sessions are recorded on the same day.
Usually there are around 30--50 people on the boulevard at the time of recordings.
The length of the boulevard ca. is 250 meters. The participants are instructed to walk back and forth once along the boulevard. They are encouraged to behave freely and naturally, with no specific task in mind, thereby exploring the surroundings at their own pace.

In the targeted roaming scenario, participants are given the task of locating specific rooms within the library.
In the first session (S1), they are instructed to locate the ``family room'' and in the second session (S2), the ``learning room''. The participants started from the library entrance and are free to choose their own navigation method (using the library plan, seeking assistance, exploring independently, or other). Both sessions required the participants to return to the starting point and both sessions are recorded on the same day.

\begin{figure}[!ht]
	\includegraphics[width=\textwidth]{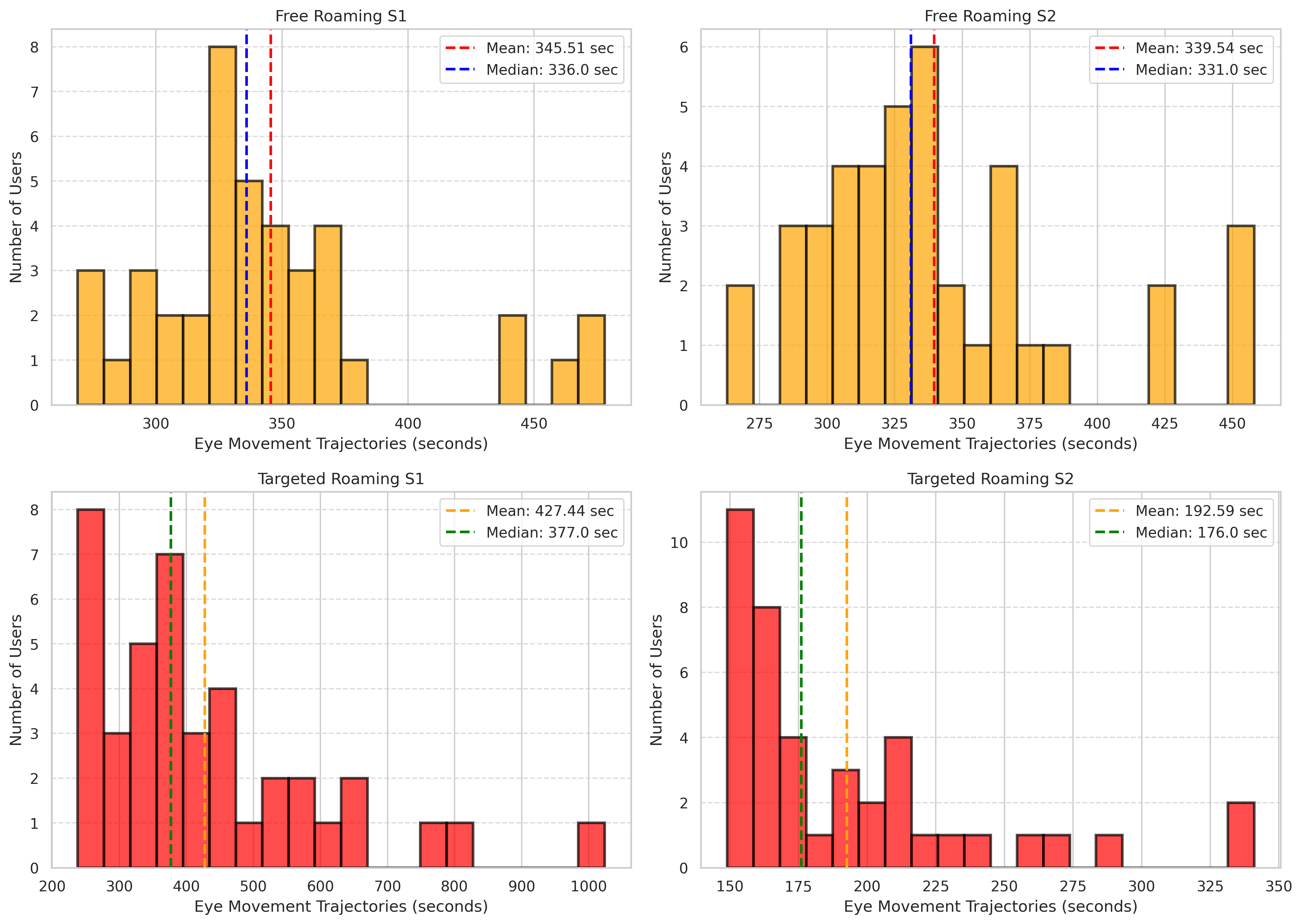}
	\caption{User distribution based on the length of their trajectories in various sessions.} \label{user_distribution}
\end{figure}

Figure~\ref{user_distribution} illustrates the distribution of trajectory lengths across the two datasets and their respective sessions. 
The shortest durations are is 263s for FR and 154s for SR.
The longest trajectory duration in FR is observed to be around 470s in S1 and 450s in S2. The durations of the trajectories in FR are similar across both sessions.
In comparison, the longest trajectory in TR is around 1000s in S1 and 340s in S2.
%This difference is due to the task in S1 requiring participants to locate a specific room on the same floor, leading them to find the learning room more easily.
One issue with the TR dataset is the layout of the rooms. The proximity of the second room to the first one makes it easier for participants to find the second room in S2. As a result, sessions for each room vary in duration, with longer sessions in S1 which is used for training and shorter sessions in S2 used for testing. These difference could reduce the accuracy of our approach.

\subsection{Data Processing and Segmentation}

We only use the ``raw'' x, y coordinate trajectory data provided by the eye tracker.
In order to reduce the noise in the raw data, we apply a Savitzky Golay filter~\cite{press1990savitzky, schafer2011savitzky}.
This filter applies a symmetric polynomial over several points. 
We use polynomial order of 6 and frame size of 15 as used in the works of~\cite{george2016score, schroder2020robustness}.

\begin{figure}[!ht]
	\includegraphics[width=\textwidth]{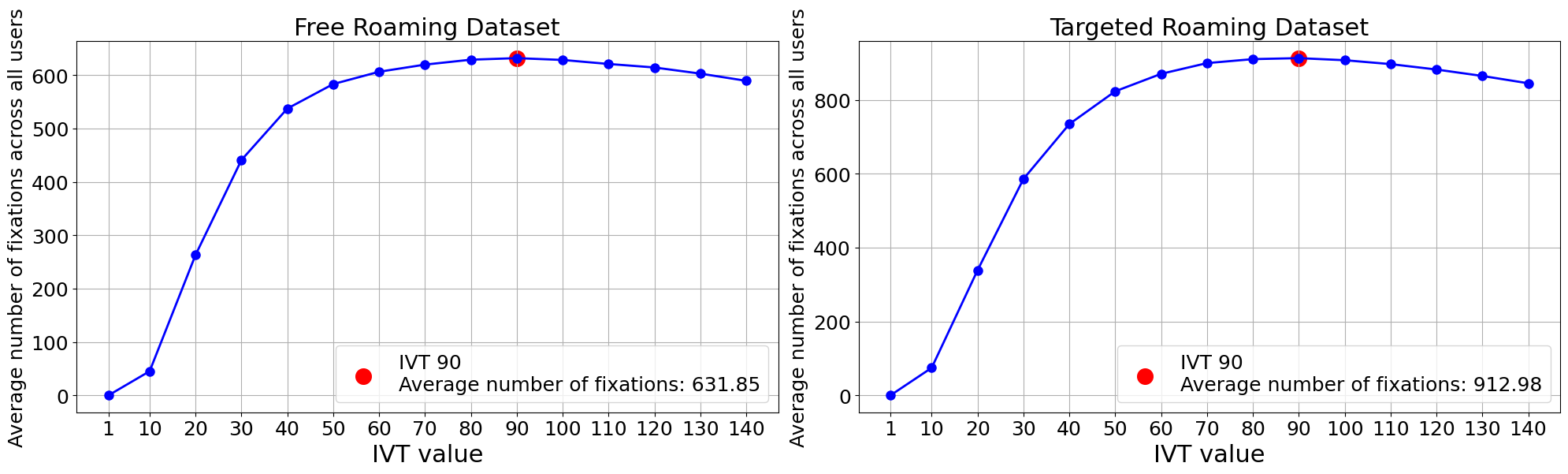}
	\caption{Average number of fixations across all users per VT value for the FR and the TR dataset.} \label{IVTT}
\end{figure}

The \emph{Identification by Velocity Threshold}(IVT) \cite{salvucci2000identifying} algorithm, as employed in various publications such as~\cite{george2016score, al2022extensive}, is used for segmenting trajectories into fixations and saccades. In our study, we adopt the IVT version used in~\cite{george2016score}.
This algorithm relies on two parameters: the \emph{velocity threshold}(VT) and the \emph{minimum fixation duration}(MFD). Segments of the gaze trajectory characterized by a velocity below the specified VT and a duration exceeding the defined MFD are classified as fixations. All other segments are classified as saccades.
The MFD value of 96ms is used~\cite{al2022extensive}.
In our segmentation method, we examined the range of VT values from 1 to 150. 
This is examined after combining both the sessions S1 and S2 for FR and TR datasets.
For each VT value, we calculate the mean number of fixations across all users. 
Figure~\ref{IVTT} illustrates the mean number of fixations across all users at different VT values on both FR and TR datasets. In order to maximize the mean number of fixations across all users, a VT of 90 deg/s is selected for both the FR and the TR datasets.
At a VT value of 90, the targeted roaming dataset exhibits an average fixation count of 913, approximately 1.5 times greater than the free roaming dataset. 
This change could be due to a higher attention in TR, because it is task oriented. 
This approach to VT parameter selection has been shown to yield high accuracies for identification tasks~\cite{al2022extensive}.

\subsection{Feature Extraction}
\label{sec_feature_extraction}

In our study, we take over the features of~\cite{al2022extensive}.
They compute fundamental eye movement features such as duration (average time duration over all fix/sac), path length (the total distance covered by the eye during a fix/sac.), fix/sac ratio (maximum angular velocity divided by the duration), fix/sac angle (angle between consecutive fix/sac), amplitude (range of fix/sac), dispersion (measures the spread or variability of fix/sac), etc. 

Following the benefits of higher order derivative features in~\cite{al2022extensive}, we take the same feature set and include derivatives up to the fifth order, computed through the forward difference method. 
Statistical features (mean, median, max, std, skewness, and kurtosis) for velocities and derivatives were derived, collectively termed ``M3S2K''.
Table~\ref{fix_sacc_features} shows the sets of features that are employed in our experiments.
We normalized all features using the Z score standardization method~\cite{pedregosa2011scikit}.

%\begin{table*}[!h]
%	\centering
%	\caption{User identification features.}
%	\label{fix_sacc_features}
%	\begingroup
%	\setlength{\tabcolsep}{6pt} % horizontal distance Default value: 6pt % vertical distance Default value: 1
%	\scalebox{1}{
%		\begin{tabular}{llllll}
%			\toprule
%			\textbf{} & \textbf{Fix./Sac. Features} & \textbf{} & \textbf{Fix./Sac. Features} & \textbf{} & \textbf{Fix./Sac. Features} \\
%			\midrule
%			1 & Duration & 11 & Amplitude & 46-51 & {\underline{Acceleration Y*}} \\
%			2 & Path length & 12 & Dispersion & 52-57 & Angular jerk* \\
%			3 & Skew X & 13 & Dist. with previous Fix/Sac & 58-63 & Jerk X* \\
%			4 & Skew Y & 14 & Angle with previous Fix/Sac & 64-69 & {\underline{Jerk Y*}} \\
%			5 & Kurt X & 15 & {\underline{Average velocity}} & 70-75 & Angular jounce* \\
%			6 & Kurt Y & 16-21 & Angular velocity* & 76-81 & Jounce X* \\
%			7 & STD of X & 22-27 & Velocity X* & 82-87 & {\underline{Jounce Y*}} \\
%			8 & STD of Y & 28-33 & {\underline{Velocity Y*}} & 88-93 & Angular crackle* \\
%			9 & Fix/Sac ratio & 34-39 & Angular acceleration* & 94-99 & Crackle X* \\
%			10 & Fix/Sac angle & 40-45 & Acceleration X*  & 100-105 & Crackle Y* \\
%			\bottomrule
%		\end{tabular}
%	}
% *M3S2K-Statistical features: Mean, Median, Max, STD, Skewness, Kurtosis
%	\endgroup 
%\end{table*}

\begin{table*}[!ht]
    \centering
    \caption{User identification fixation and saccade features.}
    \label{fix_sacc_features}
    \begingroup
    \setlength{\tabcolsep}{6pt} % horizontal distance Default value: 6pt % vertical distance Default value: 1
    \scalebox{1}{
        \begin{tabular}{clclcl}
            \toprule
            \multicolumn{6}{l}{\textbf{Position based 15 features}} \\
            \midrule
            1 & Duration & 6 & Kurtosis Y & 11 & Amplitude \\
            2 & Path length & 7 & Standard deviation of X & 12 & Dispersion \\
            3 & Skew X & 8 & Standard deviation of Y & 13 & Dist. to previous Fix/Sac \\
            4 & Skew Y & 9 & Fix/Sac ratio & 14 &Angle with previous Fix/Sac \\
            5 & Kurtosis X & 10 & Fix/Sac angle & 15 & Average speed \\
            \midrule
            \multicolumn{6}{l}{\textbf{Higher order derivative features (18 features per derivative) }} \\
            \midrule
            16--21 & Angular velocity* & 22--27 & Velocity X* & 28--33 & Velocity Y* \\  
            \midrule
            34--39 & Angular acceleration* & 40--45 & Acceleration X* & 46--51 & Acceleration Y* \\  
            \midrule
            52--57 & Angular jerk* & 58--63 & Jerk X* & 64--69 & Jerk Y* \\  
            \midrule
            70--75 & Angular jounce* & 76--81 & Jounce X* & 82--87 & Jounce Y* \\
            \midrule
            88-93 & Angular crackle & 94--99 & Crackle X* & 100-105 & Crackle Y* \\
            \bottomrule            
        \end{tabular}
    }
 *M3S2K-Statistical features: Mean, Median, Max, Standard deviation, Skewness, Kurtosis
    \endgroup 
\end{table*}

\subsection{Machine Learning Classifiers and Performance Metrics}
\label{ML_classifiers}
The Radial Basis Function Network (RBFN) classifier~\cite{broomhead1988radial, george2016score} proves to be the most effective for our study. In comparison with other classifiers such as Logistic Regression \cite{bishop2006pattern_chap4}, Random Forest \cite{breiman2001random}, Support Vector Machines \cite{cortes1995support}, and Naïve Bayes \cite{bayes1968naive}, our evaluation reveals that RBFN consistently achieves better accuracies. 
This coincides with previous findings~\cite{schroder2020robustness, al2022extensive}.

We utilize the RBFN classifier with \(k=32\), a standard choice known for balancing model performance and computational efficiency, as mentioned in~\cite{george2016score, al2022extensive}.
Two instances of RBFN classifiers are trained: one for predicting users from fixations and another for predicting users from saccades. 
For each instance, we identify the identified user among a set of 41 different users based on probability scores assigned to each user. 
To assess the \emph{accuracy} we use the formula: dividing the number of correct predictions by the total number of predictions (i.e., the number of users). 
The final prediction probability $P_{\text{final}}(i)$ is computed as the average of the probabilities obtained from the fixation $P_{\text{fix}}(i)$ and saccade $P_{\text{sac}}(i)$ classifiers for each class $i$ (userID).

\begin{equation}
	P_{\text{final}}(i) = 0.5 \cdot P_{\text{fix}}(i) + 0.5 \cdot P_{\text{sac}}(i)
\end{equation}

%\begin{equation}
%\sigma = \sqrt{\frac{\sum_{i=1}^{N} (\text{accuracy}_{\text{x}} - \mu)^2}{N}}
%\end{equation}

All our accuracies are reported as percentage points.
Given the stochastic nature of the RBFN classifier, we conduct cross validation across 50 different states (seeds). Each experiment is repeated with a unique seed to capture the variability introduced by the random initialization of the RBFN algorithm's internal state. Subsequently, we compute the average accuracy across all 50 seeds. 
Together with accuracy, we also report the \emph{standard deviation}~(SD).
We use S1 for training and S2 for testing, for both the FR and TR datasets.

\section{User Identification Experiments}
\label{sec_Experiments}
In this experiment we start with a basic approach without optimizing trajectory length. 
The work of \cite{al2022extensive} found that adding up to the fifth order of derivative features has a big impact on identification accuracy.
Following their approach, we use different sets of higher order features, beginning with 15 position based features (Table \ref{fix_sacc_features}). 
Then, we add velocity features (16--33), then acceleration (34--51), jerk (52--69), jounce (70--87), and finally crackle (88--105).
We now use the complete trajectory for FR and TR using S1 for training and S2 for testing.	
In the FR dataset (see Table~\ref*{Accuracies_FULL}), the highest accuracy is 82.2\% when incorporating Jounce higher order derivative. Conversely, for the targeted roaming dataset, the accuracy is notably lower at 37.9\% with the inclusion of Crackle higher order derivative.

%In order to study the higher order derivative features, we start with a minimal set of first 14 position based features and duration as a general feature. 
%See the first 14 features in Table~\ref{fix_sacc_features}. 
%Next we add 19 velocity based features and than, step by step, we include 18 statistical features based on each of the other higher order derivatives (acceleration, jerk, jounce, crackle).

\begin{table*}[!ht]
	\centering
	\caption{Accuracies and SD for FR, and TR dataset over 50 runs.}
	\label{Accuracies_FULL}
	\begingroup
	\setlength{\tabcolsep}{6pt} % horizontal distance Default value: 6pt % vertical distance Default value: 1
	\scalebox{1}{
		\begin{tabular}{lcll}
			\toprule
			\textbf{Higher Order} & \textbf{No of} & \textbf{FR full} & \textbf{TR full} \\
			\textbf{Derivative} & \textbf{features} & \textbf{} & \textbf{} \\
			\midrule
			Position (0) & 14 & 66.9 $\pm$ 1.9 & 22.7 $\pm$ 1.3 \\
			Velocity (1) & 33 & 79.1 $\pm$ 2.0 & 33.0 $\pm$ 1.5 \\
			Acceleration (2) & 51 & 80.9 $\pm$ 1.5 & 37.3 $\pm$ 1.1 \\
			Jerk (3) & 69 & 81.5 $\pm$ 1.6 & 37.2 $\pm$ 1.5 \\
			Jounce (4) & 87 & \textbf{82.2 $\pm$ 1.2} & 37.0 $\pm$ 2.4 \\
			Crackle (5) & 105 & 80.6 $\pm$ 1.6 & \textbf{37.9 $\pm$ 1.4} \\
			\bottomrule
		\end{tabular}
	}
	\endgroup
\end{table*}

\subsection{Trajectory Fragments}

To improve the baseline approach, we employ the following method. As in Section \ref{Baseline} we vary the number of features.
But now we additionally take only fragments of the trajectory for training (S1) and testing (S2).
For our experiments, we utilize the following fragments of the trajectory: 60s, 80s, 100s, 120s, 130s, 140s, and 150s. 
We cut these fragments either from the beginning or from the end of the trajectories for both sessions (S1, S2). 

\begin{figure}[!ht]
	\includegraphics[width=0.83\textwidth]{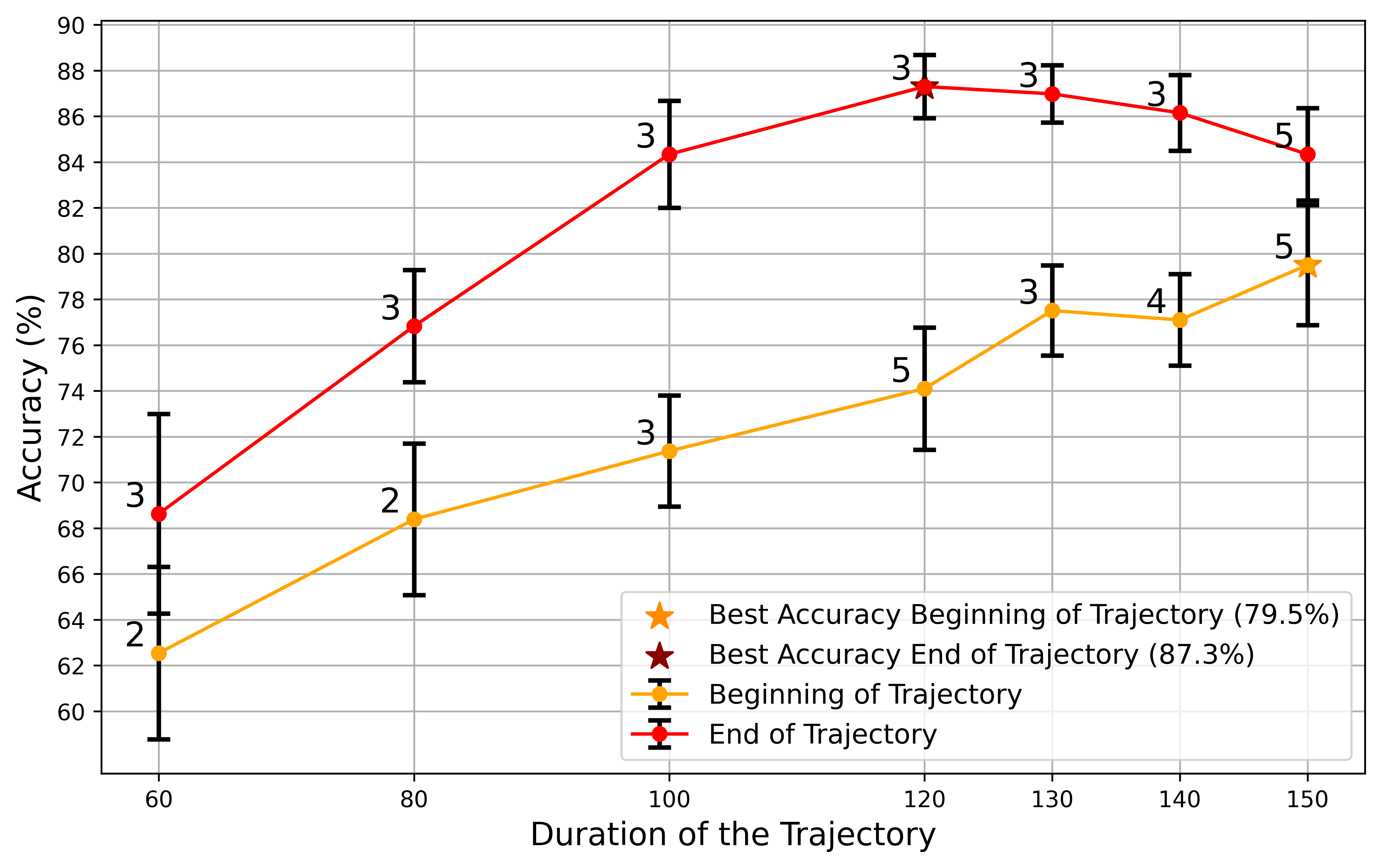}
	\caption{Accuracies and SD over fragments of trajectories from the beginning and the end for the FR data over 50 runs.} \label{Free}
\end{figure}

\begin{figure}[!htbp]
	\includegraphics[width=0.83\textwidth]{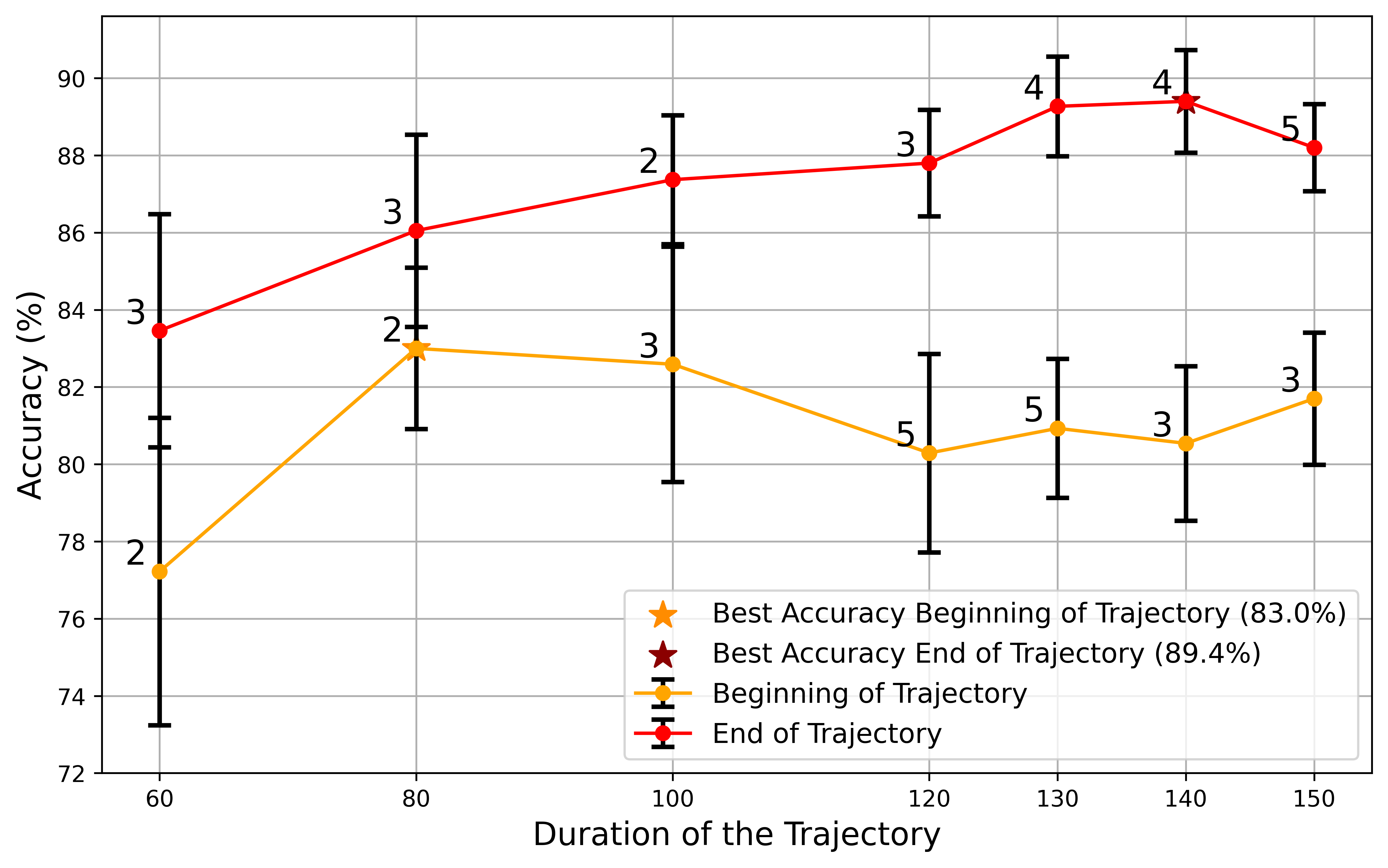}
	\caption{Accuracies and SD over fragments of trajectories from the beginning and the end for the TR data over 50 runs.} \label{targeted}
\end{figure}

We only show our best accuracies over all the feature sets in each time fragment from the start and the end for the FR and TR datasets. 
Figure~\ref{Free} shows all the best results for the FR dataset for fragments taken from beginning of the trajectory in yellow and for fragments taken from the end of the trajectory in red.
The best accuracies for the FR dataset is 87.3 $\pm$ 1.3 which is obtained from fragments of 120s  trajectory from the end. 
In the Figure~\ref{Free} and~\ref{targeted}, the label near the accuracy denotes the utilization of higher order derivative features, e.g. if 4 is the label then features upto higher order of jerk (69 features from~\ref{fix_sacc_features}) are utilized for the experiment.
Figure~\ref{targeted} shows all the best results for the TR dataset for fragments taken from start of the trajectory in yellow and for fragments taken from the end of the trajectory in red.
For the TR dataset the best accuracy is 89.4 $\pm$ 1.3 which is obtained from fragments of 140s  trajectory from the end.

%\textbf{Trajectory fragments}:
%In order to study the effect of the length of the trajectories as mentioned in~\cite{al2022extensive} we take fragments of the trajectories for each user from the start and the end. 
%The following lengths are used for our experiments 60sec, 80sec, 100sec, 120sec, 130sec, 140sec, 150 sand similarly from the end for both sessions (S1, S2) individually and for both datasets (FR, TR).

%comparison RAN FR TR
\subsection{Comparison with BioEye RAN Dataset} 
To the best of our knowledge the study of~\cite{al2022extensive} achieves the highest accuracy for user identification using eye tracking. 
The dataset stems from the~\emph{BioEye}~\cite{komogortsev2015bioeye} (TEX/RAN) competition and was recorded using an EyeLink 1000 eye tracker (1000Hz). 
The data was down sampled to 250Hz using an anti aliasing filter.
Two datasets with different stimuli are used.
TEX: 60s of recording of reading a poem on a computer screen.
RAN: 100s of recording of observing a random moving dot on a computer screen.
The study comprises 153 participants. 
The users were seated at a distance of 550mm from the computer screen.
Participants' heads were stabilized using a chin rest to minimize head movement related eye tracking artifacts. 
They~\cite{al2022extensive} report an accuracy of 96.0 $\pm$ 0.6 for RAN. 
This is obtained using RBFN classifier over 50 runs.
The experiments are performed in an controlled environment using carefully selected stimuli to extract key information from participants.

We conduct experiments to compare non laboratory FR and TR datasets with a laboratory setting RAN dataset. 
Initially, we select a subset of 41 users from the RAN dataset, achieving an accuracy of 96.2\%. 
To assess the impact of variability in eye tracking quality, we downsample the 250Hz data to 200Hz. The best accuracy reported in \cite{al2022extensive} decreases to 95.3\%. 
Another experiment investigates how downsampling affects all 153 users, reducing the accuracy by 1.2\% compared to without downsampling the data.

\begin{table*}[!ht]
	\centering
	\caption{Accuracies and SD for the RAN dataset over 50 runs using different parameters.}
	\label{RAN_153}
	\begingroup
	\setlength{\tabcolsep}{2pt} % adjust horizontal spacing
	\begin{tabular}{lcccc}
		\toprule
		\textbf{Dataset} & \textbf{153 users (250Hz)} & \textbf{41 users (250Hz)} & \textbf{41 users (200Hz)} & \textbf{153 users (200Hz)} \\
		\midrule
		Accuracy & 96.0 $\pm$ 0.6 & 96.2 $\pm$ 1.1 & 95.3 $\pm$ 1.9 & 94.7 $\pm$ 1.2 \\
		\bottomrule
	\end{tabular}
	\endgroup
\end{table*}

Table~\ref{Accuracies_RAN} shows the accuracy for the RAN, FR, and TR datasets over these setting.
We find that when taking subsets of the TR dataset from the end of the trajectory we achieve a comparable accuracy of 89.4\% to 95.3\% of RAN dataset. 
These outcomes signify a promising advancement, considering the non laboratory settings and cheap eye tracker. 

\begin{table*}[!ht]
	\centering
	\caption{Accuracies and SD for FR, TR, and RAN dataset over 50 runs.}
	\label{Accuracies_RAN}
	\begingroup
	\setlength{\tabcolsep}{6pt} % horizontal distance Default value: 6pt % vertical distance Default value: 1
	\scalebox{1}{
		\begin{tabular}{lclll}
			\toprule
			\textbf{Higher Order} & \textbf{No of} & \textbf{FR 120s} & \textbf{TR 140s} & \textbf{RAN 100s} \\
			\textbf{Derivative} & \textbf{features} & \textbf{from the end} & \textbf{from the end} & \textbf{full} \\
			\midrule
			Position (0) & 14 & 75.8 $\pm$ 3.3 & 70.1 $\pm$ 4.2 & 88.9 $\pm$ 1.8 \\
			Velocity (1) & 33 & 84.1 $\pm$ 2.5 & 85.8 $\pm$ 3.2 & \textbf{95.3 $\pm$ 1.9} \\
			Acceleration (2) & 51 & 86.9 $\pm$ 1.4 & 88.3 $\pm$ 1.1 & 95.1 $\pm$ 1.5 \\
			Jerk (3) & 69 & \textbf{87.3 $\pm$ 1.3} & 88.6 $\pm$ 1.6 & 95.2 $\pm$ 1.5 \\
			Jounce (4) & 87 & 85.3 $\pm$ 2.2 & \textbf{89.4 $\pm$ 1.3} & 95.0 $\pm$ 1.6 \\
			Crackle (5) & 105 & 84.4 $\pm$ 2.0 & 88.8 $\pm$ 1.7 & 94.9 $\pm$ 0.5 \\
			\bottomrule
		\end{tabular}
	}
	\endgroup
\end{table*}

\section{Top Features} 
\label{Statistics}
Using ANOVA~\cite{st1989analysis}, we calculate the top features for both the FR and TR datasets. 
While computing the top features we merge both sessions S1 and S2 for the FR and TR datasets.
ANOVA assesses whether group means, representing different users in this context, differ significantly by comparing the ratio of variability between these user groups. This analysis results in an F-score that indicates the extent of differences between these user groups.
Table~\ref{top_feature_stats} shows that duration is a top feature for both the FR and TR datasets. 
In the FR scenario, fixation duration may show curiosity and exploration of surroundings. In the TR scenario, extended fixations help in a focused search for specific locations, aiding in finding the room.

% \usepackage{booktabs}
\begin{table}[!ht]
	\centering
	\caption{Top features for FR and TR datasets (full trajectory) using ANOVA.}
	\label{top_feature_stats}
	\begin{tabular}{@{}lclclclc@{}}
		\toprule
		\multicolumn{4}{c}{\textbf{FR full trajectory}} & \multicolumn{4}{c}{\textbf{TR full trajectory}} \\
		\cmidrule(lr){1-4} \cmidrule(lr){5-8}
		\textbf{Fixation} & \textbf{Score} & \textbf{Saccade} & \textbf{Score} & \textbf{Fixation} & \textbf{Score} & \textbf{Saccade} & \textbf{Score} \\ 
		\midrule
		 Duration            & 6.69           &  Duration            & 14.07          &  Duration            & 5.98           &  Duration            & 11.90           \\
		Std ang acceleration        & 3.74           & Mean ang velocity        & 2.47           & Std acceleration Y           & 2.85           & Mean ang velocity        & 1.14            \\
		Std acceleration X          & 3.10           & Avg velocity             & 2.27           & Std ang acceleration         & 2.83           & Avg velocity             & 1.10            \\
		Std velocity X          & 2.62           & Median ang velocity      & 2.08           & Std acceleration X           & 2.78           & Median ang velocity      & 1.00            \\
		Std acceleration Y          & 2.35           & Sac ratio          & 1.66           & Std velocity Y           & 2.39           & Skew X              & 0.98            \\ 
		\bottomrule
	\end{tabular}
\end{table}

We compare the top features between full trajectory and best fragments of FR and TR datasets. 
The best fragments are 120s from the end for the FR and 140s from the end for the TR.
Primary features remained consistent, but the fifth top feature for fixation classifier differed in both the datasets.
ANOVA F-scores for duration indicate significant variability across users and datasets. Further investigation is needed to understand top feature contributions to accuracy. Overall, top features are similar and stable across the FR and TR datasets.

%% Please add the following required packages to your document preamble:
%% \usepackage{booktabs}
%\begin{table}[!ht]
%	\centering
%	\caption{Top features of fixation and saccade classifier over best time fragments for FR (120s from the end) and TR (140s from the end) datasets using ANOVA.}
%	\label{top_feature_stats_end}
%	\begin{tabular}{@{}lclclclc@{}}
%	\toprule
%	\multicolumn{4}{c}{\textbf{FR 120s from the end}} & \multicolumn{4}{c}{\textbf{TR 140s from the end}} \\
%	\cmidrule(lr){1-4} \cmidrule(lr){5-8}
%	\textbf{Fixation} & \textbf{Score} & \textbf{Saccade} & \textbf{Score} & \textbf{Fixation} & \textbf{Score} & \textbf{Saccade} & \textbf{Score} \\ 
%	\midrule
%	Duration            & 17.19           &  Duration            & 8.34          &  Duration            & 11.64           &  Duration            & 10.32           \\
%	Std ang acceleration        & 1.62           & Mean ang velocity        & 6.65           & Std acceleration Y           & 1.21           & Mean ang velocity        & 4.94            \\
%	Std acceleration X          & 1.52           & Avg velocity             & 1.01           & Std ang acceleration         & 1..20           & Avg velocity             & 4.75            \\
%	Std velocity X          & 1.26           & Median ang velocity      & 0.97           & Std acceleration X           & 1.18           & Median ang velocity      & 4.43            \\
%	Mean ang velocity          & 0.95           & Sac ratio          & 0.77           & Max acceleration Y           & 0.79           & Skew X              & 4.41            \\ 
%	\bottomrule
%\end{tabular}
%\end{table}

\begin{figure}[!ht]
	\includegraphics[width=\textwidth]{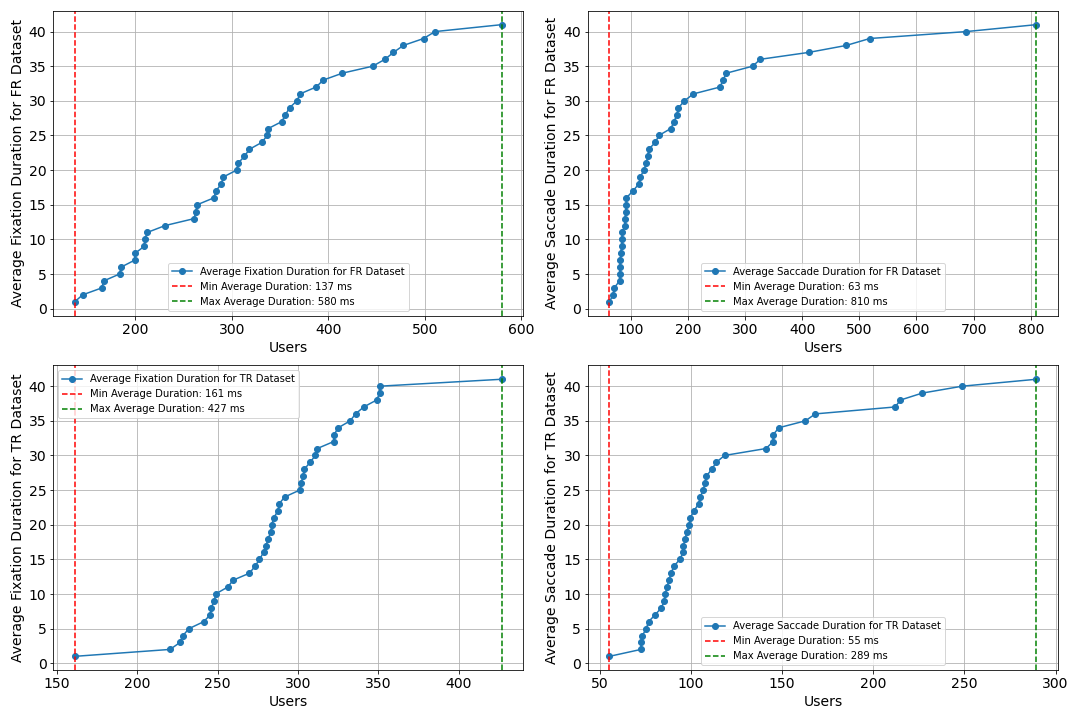}
	\caption{Average fixation and saccade duration over all the users for both the FR and TR datasets.} \label{Avg}
\end{figure}

Considering that duration stands out as the top feature, we examine the average duration values for both fixation and saccades in both datasets. Figure~\ref{Avg} illustrates the distribution. It is evident that fixations are fairly evenly distributed, facilitating easier differentiation. In contrast, saccades tend to be skewed towards lower values, although some users exhibit exceptionally high saccade durations.

\section{Conclusion}
\label{Conclusion}
In conclusion, our study presents comparable results for user identification using free roaming eye movements and a low cost commercial 200Hz eye tracker achieving an accuracy of 89.4\%. Future research should focus on achieving task independence by training on diverse datasets and testing on new ones. 
Additionally, exploring the optimal trajectory length and increasing the user sample size will contribute to the robustness of the user identification model.

\bibliographystyle{splncs04}
\bibliography{content/HCI_2024}

\end{document}